\documentclass[11pt]{article}
\usepackage{arxiv}

\usepackage{latexsym}
\usepackage[T1]{fontenc}
\usepackage[utf8]{inputenc}
\usepackage{graphicx}
\usepackage{caption}
\usepackage{comment}
\usepackage{tabularx}
\usepackage{longtable}
\usepackage[numbers]{natbib}
\usepackage[unicode=true,colorlinks]{hyperref}
\hypersetup{
    pdftitle={Synthetic Data Generation Using Large Language Models: Advances in Text and Code},
    pdfauthor={Mihai Nadăș, Laura Dioșan, Andreea Tomescu},
    pdfkeywords={Synthetic Data Generation, Large Language Models, Text Data Augmentation, Code Data Synthesis, Prompt Engineering, Instruction Tuning, Machine Learning Training Data, Natural Language Processing, Code Generation, Reinforcement Learning for Code, Automated Data Annotation, Bias and Fairness in Synthetic Data, Retrieval-Augmented Generation, Evaluation of Synthetic Data, Model Collapse in LLMs},
    bookmarksnumbered,
    pdfstartview={FitH},
    urlcolor=cyan,
    colorlinks=true,
    linkcolor=red,
    citecolor=green,
}

\title{Synthetic Data Generation Using Large Language Models: Advances in Text and Code}

\author{
  Mihai Nad\u{a}\c{s}$^{1}$ \and
  Laura Dio\c{s}an$^{1}$ \and
  Andreea Tomescu$^{2}$ \\
  $^{1}$Faculty of Mathematics and Computer Science, Babe\c{s}-Bolyai University, Cluj-Napoca, Romania \\
  $^{2}$KlusAI Research Lab, Cluj-Napoca, Romania \\
  \texttt{\{mihai.nadas, laura.diosan \}@ubbcluj.ro} \\
  \texttt{\{andreea.tomescu\}@klusai.com}
}

\begin{document}

\maketitle

\begin{abstract}
This survey reviews how large language models (LLMs) are transforming synthetic training data generation in both natural language and code domains. By producing artificial but task-relevant examples, these models can significantly augment or even substitute for real-world datasets, particularly in scenarios where labeled data is scarce, expensive, or sensitive. This paper surveys recent advances in leveraging LLMs to create synthetic text and code, highlighting key techniques such as prompt-based generation, retrieval-augmented pipelines, and iterative self-refinement. We examine how these methods can enrich low-resource tasks (e.g., classification, question answering) and facilitate code-centric applications (e.g., instruction tuning, code translation, bug repair) through automated verification of functional correctness. Alongside potential benefits---cost-effectiveness, broad coverage, and controllable diversity---we discuss the accompanying challenges, including factual inaccuracies in generated text, insufficient stylistic or distributional realism, and risks of bias amplification. Proposed mitigation strategies range from filtering and weighting synthetic outputs to reinforcement learning with execution feedback in code domains. We conclude by outlining open research directions, such as automated prompt engineering, cross-modal data synthesis, and robust evaluation frameworks, underscoring the growing importance of LLM-generated synthetic data in accelerating AI development while emphasizing ethical and quality safeguards.
\end{abstract}


\section{Introduction}
\label{sec:introduction}


Large Language Models (LLMs) have achieved remarkable success across natural language and code generation tasks, owing in part to training on massive datasets. However, acquiring sufficient high-quality training data remains a bottleneck in many domains \cite{yu_what_2024}. Data scarcity, high annotation costs, and privacy constraints often limit the availability of large supervised corpora. These challenges have spurred growing interest in synthetic data generation, where additional training examples are produced artificially rather than collected from the real world. Recent advances in generative AI – particularly LLMs like Anthropic's Claude 3.7 Sonnet, DeepSeek's R1, Meta's Llama 3, and OpenAI's GPT-o3 – provide powerful new tools to generate synthetic text and code that mimic real data distributions. This paper surveys and analyzes the latest developments in LLM-driven synthetic data generation for both natural language text and programming code, highlighting techniques, applications, challenges, and future directions.

LLMs can produce human-like text and code, making them attractive data generators for tasks where obtaining real data is costly or infeasible. For example, instead of manually labeling thousands of sentences for a classifier, one can prompt an LLM to create diverse labeled examples \cite{li_data_2024}, or have it generate code snippets to augment code model training. Synthetic data generated by LLMs has shown promise in boosting model performance in low-resource settings, reducing annotation costs, and enabling data augmentation for improved robustness \cite{ding_is_2023}.
At the same time, using LLMs as data generators raises questions about quality control, realism, and biases in the generated data. 

A comprehensive review of LLM-based approaches for text and code generation is currently lacking, despite their increasing adoption \cite{wang_survey_2024}. This paper aims to fill that gap by providing a structured survey of existing techniques and challenges. The focus is specifically on text and code generation, as opposed to other forms of data, due to several key reasons. First, visual data synthesis has been extensively studied, making an additional review redundant \cite{dhariwal_diffusion_2021, couairon_diffedit_2022}. Second, while synthetic signal data (e.g., audio) is a developing field, it remains outside the scope of this study, though future work could extend these analyses \cite{huang_datagen_2024}. Lastly, text and code share a common modality, as both are purely symbolic representations that can be generated and processed using the same fundamental methods—unlike images, sounds, or videos, which require distinct encoding formats and specialized equipment \cite{lachaux_unsupervised_2020}.

This paper provides a comprehensive analysis of LLM-based synthetic data generation, with contributions summarized as follows:

\begin{itemize}
    \item \textbf{Survey of Techniques}: We review major approaches for LLM-driven data generation in text and code, including prompt-based augmentation, retrieval-augmented generation, self-instruct methods, and reinforcement learning with feedback. We categorize methods by their strategies (e.g., zero-shot vs. few-shot prompting, knowledge integration, iterative refinement) and discuss representative examples from recent literature \cite{chai_text_2025}

    \item \textbf{Advances in Text Data Generation}: We explore how LLMs are used to create synthetic text datasets for tasks such as classification, question answering, and instruction-following. Key studies are examined to understand the impact of synthetic data on model performance, data diversity, and efficiency. We highlight empirical findings (e.g., improvements of 3–26\% with synthetic augmentation in low-data regimes) and analyze how prompt design and data curation affect quality \cite{li_data_2024}.
    
    \item \textbf{Advances in Code Data Generation}: We survey the parallel emergence of LLM-generated code data for training code intelligence models. We discuss techniques for generating code through LLM prompts, executing code to validate correctness, and synthesizing coding instruction data at scale \cite{liu_best_2024}. Examples include synthetic programming problems and solutions, code edits for diversity, and instruction-tuning datasets (e.g., Code Alpaca, WizardCoder) generated by LLMs.
    
    \item \textbf{Challenges and Mitigations}: We outline the main challenges in using LLMs for synthetic data, such as ensuring data fidelity and factuality, avoiding modeling biases, maintaining diversity without drifting from real data distribution, and preventing \textit{model collapse} \footnote{Model collapse refers to the degradation of model quality that occurs when successive generations of models are trained primarily or exclusively on synthetic data produced by other models, leading to loss of diversity, factuality, or robustness; see~\cite{gerstgrasser_is_2024}.} from iterative self-training on AI-generated data. We discuss solutions proposed in literature, like filtering and weighting synthetic examples \cite{li_data_2024}, blending synthetic and real data \cite{gerstgrasser_is_2024}, and leveraging feedback (e.g., executors for code, retrieval for facts) to improve quality \cite{gehring_rlef_2025} \cite{khaliq_ragar_2024}.
    
    \item \textbf{Future Directions}: We identify open research directions, including more rigorous evaluation metrics for synthetic data quality, techniques for controlling generation (to target rare phenomena or reduce biases), expanding synthetic data generation to multimodal and low-resource language settings, and frameworks for safe and privacy-preserving synthetic data in industry applications. Throughout, we connect recent findings to prior work to provide a well-rounded perspective on how LLM-based synthetic data generation is shaping the future of NLP and software engineering research.
\end{itemize}

To guide readers through the wide landscape of LLM-driven synthetic data generation, Table~\ref{tab:taxonomy} presents a high-level taxonomy of tasks addressed in the literature. We distinguish between applications in the text and code domains, further categorizing major synthetic data tasks such as generation, translation, instruction tuning, and repair. This organizational scheme forms the backbone of our survey and is reflected in the structure of the following sections.

\begin{table*}[!htbp]
\centering
\caption{Taxonomy of LLM-based synthetic data tasks}
\label{tab:taxonomy}
\begin{tabular}{|l|l|}
\hline
\textbf{Text Domain}                 & \textbf{Code Domain} \\
\hline
Generation (augmentation, prompts)   & Generation (code synthesis) \\
Translation (cross-lingual)          & Translation (code-to-code) \\
Instruction Tuning (Alpaca, etc.)    & Instruction Tuning (Code Alpaca, WizardCoder) \\
Paraphrasing / Style Transfer        & Refactoring / Style Transfer \\
QA / Reasoning (chain-of-thought)    & QA / Problem Generation \\
Retrieval-Augmented Generation       & Retrieval-Augmented Generation \\
Evaluation \& Filtering              & Evaluation (execution feedback) \\
Repair (N/A)                         & Repair (bug fixing, code improvement) \\
\hline
\end{tabular}
\end{table*}

The rest of this paper is organized as follows. Section 2 provides background on synthetic data and early data augmentation approaches. Section 3 and 4 detail advances in synthetic data generation for text and code, respectively. Section 5 discusses overarching challenges and comparative insights. Section 6 suggests promising future research directions, and Section 7 concludes the paper.

\section{Related Work}

The field of synthetic data generation, especially with large language models (LLMs), has evolved rapidly, prompting the publication of several recent survey articles that address various aspects of this topic. We briefly review and position our work with respect to these prior surveys.

\textbf{Synthetic Data for NLP.}  
Wang et al. \cite{wang_survey_2024} provide a comprehensive overview of data synthesis and augmentation techniques for large language models in natural language processing (NLP). Their survey covers a broad spectrum of approaches, from traditional text augmentation methods to modern neural and prompt-based strategies. However, their primary focus is on NLP applications, with only limited discussion of synthetic data for programming code or the distinctive challenges faced in code generation domains.

\textbf{LLM-Driven Synthetic Data Pipelines.}  
Long et al. \cite{long_llms-driven_2024} propose a generic framework for LLM-driven synthetic data generation, curation, and evaluation. Their review encompasses text, code, and other modalities, organizing the landscape around lifecycle stages such as data generation, filtering, and integration. While thorough in its coverage, the survey remains high-level, and does not systematically analyze or compare empirical findings for text and code tasks separately. Key challenges unique to code, such as leveraging execution feedback \footnote{Execution feedback uses the outcome of running generated code—such as passing or failing test cases—as a supervisory signal for selecting or further training models~\cite{li_competition-level_2022}.} and handling functional correctness, are not addressed in detail.

\textbf{Surveys on Synthetic Code Data.}  
A number of surveys focus specifically on code generation with LLMs. Jiang et al. \cite{jiang_survey_2024} review architectural advances and applications of large language models for code, with a primary emphasis on model capabilities and benchmark results. Chen et al. \cite{chen_mastering_2025} present a rich taxonomy of synthetic data generation techniques for code LLMs, cataloging a wide array of recent approaches for code instruction tuning, translation, and program repair. While these works offer valuable overviews, they are largely centered on code, with less attention given to synthetic data issues shared across natural language and code domains, such as distributional alignment, bias amplification, or unified evaluation strategies.

\textbf{Position and Novelty of This Survey.}  
In contrast to these prior works, our survey aims to bridge the gap between synthetic data generation for natural language and programming code, providing a unified and comparative treatment of LLM-driven methods across both modalities. Specifically, our contributions are:
\begin{itemize}
    \item \textbf{Unified Framework:} We synthesize and extend recent advances in both text and code synthetic data generation, providing a cross-domain taxonomy of LLM-driven techniques---including prompt-based, retrieval-augmented, and iterative self-refinement pipelines.
    \item \textbf{Comparative Analysis:} We systematically compare empirical results, challenges, and mitigation strategies in both domains, with particular focus on issues of quality assurance, distributional realism, bias, and model collapse, which are often addressed separately in prior surveys.
    \item \textbf{Practical Lessons and Future Directions:} We distill practical recommendations from recent studies and outline a forward-looking research agenda that incorporates trends in evaluation, human-in-the-loop pipelines, and ethical considerations relevant to both NLP and code generation communities.
\end{itemize}

Thus, this survey advances the literature by offering greater scope—through the joint treatment of text and code tasks---greater depth via comparative empirical analysis, and a novel structure that unifies recent methods, challenges, and practical insights across modalities.

\section{Survey Methodology}
\label{sec:methodology}

\begin{figure}[!ht]
\centering
\includegraphics[width=0.48\textwidth]{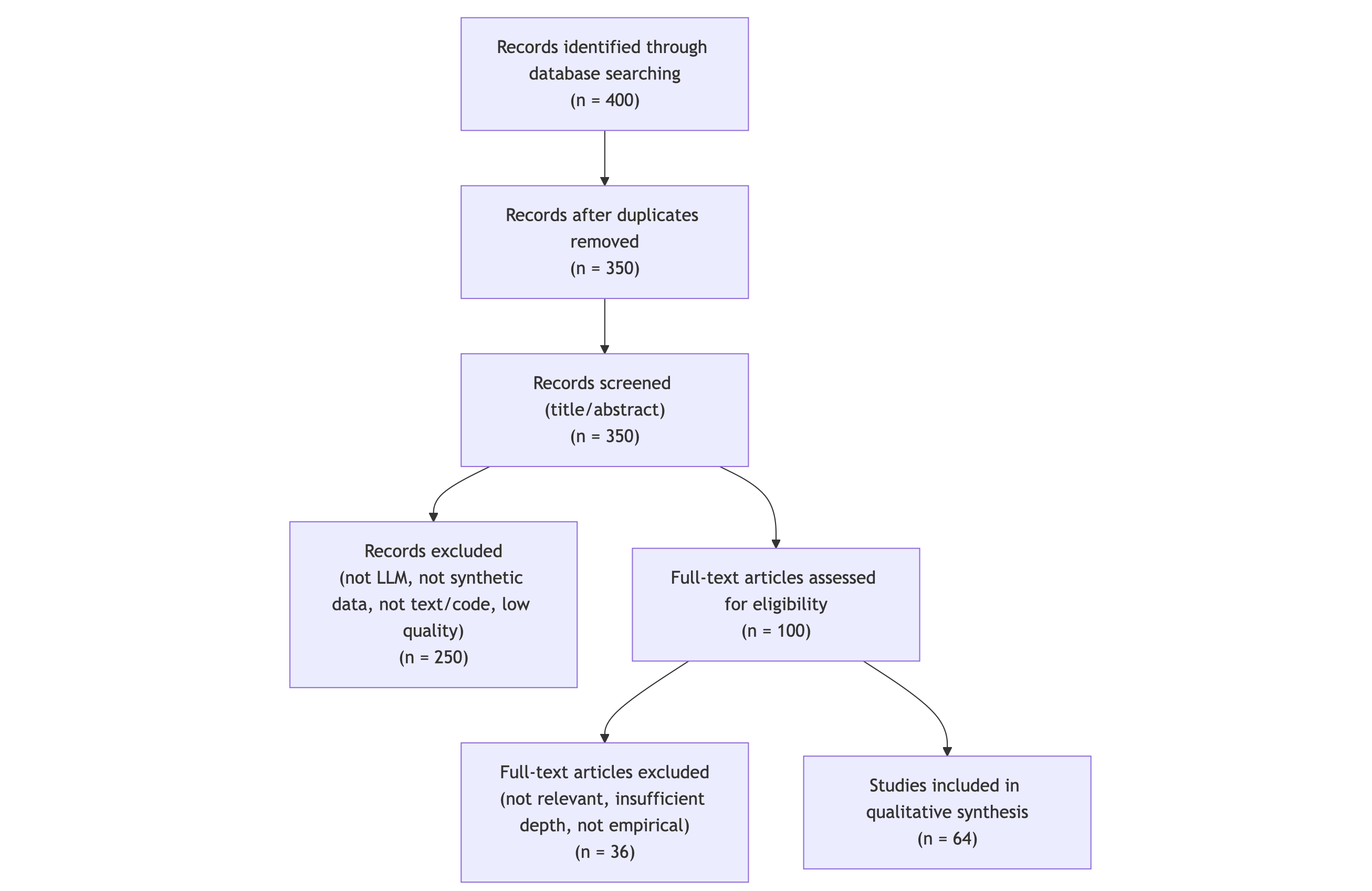}
\caption{PRISMA-style flow diagram for study selection.}
\label{fig:prisma}
\end{figure}

To ensure transparency and reproducibility, we detail here the systematic methodology employed for selecting and reviewing the literature included in this survey, following a PRISMA-style approach as summarized in Figure~\ref{fig:prisma}.

\textbf{Inclusion Criteria:}  
We considered papers and preprints that met the following criteria:
\begin{itemize}
    \item Published between January 2020 and April 2025, to capture the period of rapid advancement in LLM-driven synthetic data generation.
    \item Address synthetic data generation, data augmentation, or instruction tuning in the context of large language models (LLMs), for either text or code domains.
    \item Focus on models with at least several hundred million parameters (to ensure relevance to the LLM paradigm), including both proprietary (e.g., GPT-3/4, Claude, Gemini) and open-source (e.g., Llama, CodeLlama, StarCoder) models.
    \item Cover a range of task categories, including but not limited to: classification, question answering, instruction following, code synthesis, code translation, and bug repair.
    \item Provide empirical evidence, systematic evaluation, or substantial methodological innovation relevant to synthetic data workflows.
\end{itemize}

\textbf{Exclusion Criteria:}
\begin{itemize}
    \item Works focused exclusively on synthetic data for modalities outside text or code (e.g., images, speech) unless providing transferable methodology or insights.
    \item Non-peer-reviewed content lacking methodological rigor or empirical validation (e.g., blog posts, opinion pieces).
    \item Papers predating 2020 or focused on pre-LLM generative models, unless foundational to understanding current techniques (e.g., early data augmentation methods).
\end{itemize}

\textbf{Review Process:}  
The literature review process was conducted in a semi-automated manner. We performed systematic keyword searches on Google Scholar, arXiv, and Semantic Scholar using terms such as ``synthetic data,'' ``LLM augmentation,'' ``instruction tuning,'' ``prompt engineering,'' ``code generation,'' and related task names (e.g., ``text classification,'' ``code translation''). Reference lists of recent surveys and benchmark papers were also examined to identify additional relevant studies (snowballing). Titles and abstracts were screened to exclude irrelevant works, and full texts were reviewed for eligibility based on the inclusion/exclusion criteria above. The final set of papers reflects a curated sample that spans both influential early works and state-of-the-art studies published as of April 2025.

\textbf{Literature Screening Flow:}  
To increase transparency, Figure~\ref{fig:prisma} summarizes the identification, screening, and selection process for works included in this review, following PRISMA guidelines.

\section{Background and Motivation}

\subsection{Synthetic Data}

\textbf{Synthetic Data} refers to data that is artificially generated rather than directly collected from real-world events or annotations. In the context of AI/ML, synthetic data is created via algorithms or generative models (such as neural networks) to \textbf{mimic the characteristics} of real data \cite{liu_best_2024}. 

Early uses of synthetic data in machine learning date back decades (e.g., artificially generated images or simulated data for computer vision and robotics) \cite{sennrich_improving_2016}. In Natural Language Processing, traditional \textit{data augmentation} techniques like synonym replacement, random insertion/deletion, and back-translation have long been used to expand text training sets. For example, in machine translation, \textbf{back-translation} \cite{sennrich_improving_2016} generates synthetic parallel sentences by translating monolingual target-language data back into the source language, which significantly improved translation quality. Similarly, for text classification, methods like EDA (Easy Data Augmentation)
\cite{wei_eda_2019} applied random word-level transformations to create variant training sentences, helping reduce overfitting on small datasets. These early augmentation approaches demonstrated that even modest synthetic perturbations can improve model robustness and performance when real data is limited \cite{xie_unsupervised_2020}.

However, simple rule-based augmentations have limitations. They often do not introduce truly new linguistic patterns or semantic variations, and thus only provide limited diversity. As models grew in size and tasks became more complex, the need for more \textbf{diverse and realistic} synthetic data became evident \cite{yu_large_2023}. This paved the way for using \textbf{generative models} to create synthetic data.

Early neural approaches included using sequence-to-sequence models or language models to paraphrase text or generate new examples. For instance, unsupervised data augmentation (UDA) techniques combined back-translation with consistency training \cite{xie_unsupervised_2020} to leverage unlabeled data, and knowledge distillation methods had high-capacity teacher models label unlabeled examples for student training (a form of synthetic labeling)\cite{hinton_distilling_2015}.
These efforts hinted at the potential of \textbf{model-generated data} to stand in for human annotations.

The advent of large pre-trained language models brought synthetic data generation to a new level.

\textbf{Large Language Models} (LLMs) (with billions of parameters, trained on vast corpora) are capable of producing high-quality, coherent text that often closely resembles human-written text \cite{zubiaga_natural_2024}. Importantly, LLMs can be guided via prompts to generate data for specific tasks. For example, given a task description and a few examples (\textit{few-shot prompting}), an LLM can generate additional plausible examples for that task \cite{li_data_2024}. Even without any examples (\textit{zero-shot}), a well-crafted prompt can elicit relevant outputs for many tasks. This makes LLMs extremely flexible data generators, effectively serving as \textbf{universal data augmenters} that can create labeled data for a wide range of problems on demand \cite{yu_large_2023}.

Several factors motivate the use of LLMs for synthetic data generation:
\begin{itemize}
    \item \textbf{Data Scarcity and Cost:} In many NLP tasks (and domains like healthcare or finance), annotated data is scarce or expensive to obtain \cite{liu_best_2024}. LLMs offer a cheap and fast alternative to manual labeling. For instance, labeling 3,000 sentences for sentiment analysis was estimated to cost hundreds of dollars and many hours of human work, whereas GPT-3 could generate a similar number of examples in minutes for a fraction of the cost \cite{li_data_2024}. 
    Table \ref{tab:cost_time_comparison} 
    illustrates this cost and time disparity, as reported by \cite{ding_is_2023} for the SST-2 sentiment task.
    The Stanford Sentiment Treebank (SST) \cite{socher_recursive_2013} is a widely used dataset in Natural Language Processing (NLP), particularly for sentiment analysis and text classification. It consists of sentences extracted from movie reviews, annotated with sentiment labels at different levels of granularity.

    The SST-2 variant is a binary classification task where each sentence is labeled as either positive or negative sentiment. It is often used to train and evaluate sentiment analysis models, testing their ability to correctly classify text based on emotional tone. Given its structured annotations and real-world applicability, SST-2 serves as a benchmark for evaluating machine learning models' text classification capabilities.
    
    \item \textbf{Scalability:} Synthetic data can be generated at virtually unlimited scale. Once an effective prompt or generation pipeline is established, one can synthesize thousands or millions of examples to bolster training. This ability to \textbf{mass-produce data} is valuable for training data-hungry models and for creating balanced datasets (e.g., generating more minority class examples to address class imbalance). Researchers predict that real text data available for training may plateau in coming decades \cite{long_llms-driven_2024}, in which case synthetic data could play a crucial role in sustaining model improvements.
    
    \item \textbf{Controllability:} Unlike collected data, synthetic data generation can be tailored to specific needs. By adjusting prompts or generation criteria, one can target particular \textbf{attributes} in the data. For example, one can ensure coverage of edge cases, generate counterfactual examples to debias a model, or produce data in a specific style or reading level. Synthetic data thus provides \textbf{fine-grained control} over dataset composition (e.g., augmenting low-resource languages or rarely seen contexts) \cite{liu_best_2024}. This is difficult to achieve with natural data collection alone.
    
    \item \textbf{Privacy Preservation:} In scenarios where real data is sensitive (user data, medical records, etc.), sharing or using it directly raises privacy concerns. Synthetic data that captures aggregate patterns of the sensitive data without revealing individual-specific information can serve as a privacy-friendly alternative. For instance, an LLM could be used to generate clinical notes that are stylistically and medically similar to real notes but contain no real patient information \cite{liu_best_2024}. This enables research and model training in domains with stringent privacy requirements, effectively \textbf{anonymizing} data via generation.
    
    \item \textbf{Diversity and Robustness:} LLMs can introduce variations that humans might not think of, potentially surfacing novel phrasings or logic. This can lead to \textbf{more robust models}. By training on a wide range of synthetic scenarios (including adversarial or uncommon cases), models may generalize better and be less prone to overfitting idiosyncrasies of a small real dataset \cite{li_data_2024}. For example, generating paraphrases or semantically equivalent rephrasings of existing sentences can teach a model to handle linguistic variability.
\end{itemize}

\begin{table*}[t]
    \centering
    \caption{Cost and time comparison: Human annotation vs LLM-generated data (for 3,000 samples in SST-2 sentiment task). Synthetic data generation by GPT-3 dramatically reduces both the time and monetary cost, though resulting model accuracy is slightly lower than using human-curated data. \cite{ding_is_2023}}
    \begin{tabular}{|l|c|c|c|c|}
        \hline
        \textbf{Data Source} & \textbf{Samples (SST-2)} & \textbf{Approx. Cost} & \textbf{Time Required} & \textbf{Model Accuracy (on SST-2)} \\
        \hline
        Human-Labeled Data & 3,000 & \$221--\$300 & $\sim$1000 minutes & 88\% \\
        GPT-3 (Synthetic) & 3,000 & \$14.37 & $\sim$46 minutes & (n/a for 3k) \\
        GPT-3 (Synthetic) & 6,000 & \$28.74 (est.) & $\sim$92 minutes & 76\% \\
        \hline
    \end{tabular}
    \label{tab:cost_time_comparison}
    
    \textit{Note:} With 6,000 GPT-3 generated samples, a classifier achieved 76\% accuracy, compared to 88\% with 3,000 human-labeled samples. This highlights a quality gap, but also the improvement gained by doubling synthetic data volume.
\end{table*}

\subsection{Prompt-Based Data Augmentation with LLMs}
\label{sec:prompt-based-data-augmentation}

{Prompting techniques are at the core of LLM-driven text data generation. By providing an LLM with a carefully designed prompt, possibly including instructions and examples, one can guide it to produce new data instances for a target task \cite{li_data_2024}. Several prompting strategies exist:}

\begin{itemize}
    \item {\textbf{Zero-Shot Generation:} The prompt contains only a task instruction or description, and the LLM must generate an example from scratch
    \cite{zubiaga_natural_2024}. For instance, the instruction \textit{"Generate a new sentence expressing positive sentiment about a product"} demonstrates this method. Zero-shot prompts rely on the model's general knowledge and often yield reasonable but sometimes generic outputs. Zero-shot generation is straightforward but may produce limited diversity if used repeatedly with the same prompt.}

    \item {\textbf{One-Shot Generation:} The prompt provides the task description plus one example (input-output pair) to illustrate the format. The LLM then creates a similar example. This single demonstration can anchor the style of outputs (closer to the given example), which can improve relevance at the cost of some diversity \cite{du_glam_2022}.}

   \item {\textbf{Few-Shot Generation:} The prompt includes multiple examples (e.g., 3--5) of the task. The LLM then generates additional examples in line with the given ones. Few-shot prompting often yields higher-quality, task-specific outputs since the model can mimic patterns from the examples, making it particularly useful when the task has a specific structure or label space \cite{min_rethinking_2022}. However, as with one-shot, there is a risk of \textbf{repeating patterns} from the limited context, reducing diversity.}

    \item {\textbf{Topic/Controlled Generation:} An advanced variant introduced by Yu et al. \cite{yu_regen_2023} is \textit{zero-shot topic generation}. Here, one first uses the LLM to generate a list of relevant topics or scenarios for the task, then prompts the LLM with a selected topic to generate a specific example. This approach injects controlled diversity: each generation focuses on a different topic, leading to a more varied dataset. For instance, for sentiment classification, the LLM might generate topics like ``restaurant review'', ``electronics product review'', ``movie review'', etc., then produce a synthetic example for each. This was found to increase the diversity of synthetic datasets and often improved model training results.}

\end{itemize}

In summary, the use of LLMs for data generation is driven by the promise of \textbf{cheaper, faster, and more flexible data creation}. Nonetheless, it comes with the responsibility of ensuring that this synthetic data is as \textbf{effective and reliable} as real data. The following sections explore how researchers are harnessing LLMs to generate synthetic text and code, and what strategies are used to maximize utility while mitigating downsides.

\section{Synthetic Data Generation for Text Tasks}
\label{sec:text}

Using LLMs to generate synthetic text data has become increasingly popular for tasks like text classification, question answering, summarization, and dialog systems. We discuss prominent approaches and findings in LLM-based text data generation, from simple prompt-driven augmentation to complex pipelines that ensure quality and diversity. We also incorporate insights from recent empirical studies that evaluate the impact of synthetic data on downstream task performance.

\subsection{Various Approaches for Prompt-Based Synthetic Text Data Generation}
\label{sec:prompt-based-text-approaches}

Prompt-based synthetic data generation, discussed in detail in Section~\ref{sec:prompt-based-data-augmentation}, forms the backbone of modern LLM-driven augmentation in NLP. As previously outlined, the choice of prompting strategy—zero-shot, few-shot, or topic-controlled—critically impacts both the diversity and relevance of synthetic examples, with each approach presenting distinct trade-offs.

Recent empirical studies have systematically benchmarked these strategies. For instance, Li et al.~\cite{li_synthetic_2023} showed that, in low-resource settings, augmenting 100 real training samples with 100 GPT-3.5-generated synthetic samples can yield 3--26\% improvements in accuracy or F1 across a range of text classification tasks. The benefit was most pronounced when data scarcity led to severe model underfitting, as also observed by Chai et al.~\cite{chai_text_2025}. However, as the quantity of real data increases, the marginal advantage of synthetic augmentation diminishes, with some studies reporting negligible improvements when synthetic data is added to large real datasets~\cite{yu_what_2024}. This trend suggests that synthetic data is especially valuable as a supplement in low-data regimes, rather than a full replacement for extensive, high-quality real corpora~\cite{trabucco_effective_2023,long_llms-driven_2024,yu_large_2023}.

Beyond basic prompting, diverse strategies have emerged to further improve data utility and model robustness:
\begin{itemize}
    \item \textbf{Topic-controlled or randomized prompting} (e.g., WANLI~\cite{liu_wanli_2022}, AugGPT~\cite{dai_auggpt_2023}) systematically varies prompt content to maximize diversity across subtopics, which correlates with improved generalization—especially when training solely on synthetic data~\cite{long_llms-driven_2024, yu_regen_2023}.
    \item \textbf{Few-shot and instruction-based prompting} are preferred where format precision or specific label alignment are required, as in GPT3Mix~\cite{yoo_gpt3mix_2021} and Unnatural Instructions~\cite{honovich_unnatural_2023}. These approaches can yield highly relevant and accurate examples, even if diversity is more constrained.
    \item \textbf{Iterative and feedback-driven generation} (e.g., Self-Instruct~\cite{wang_self-instruct_2023}, SunGen~\cite{gao_self-guided_2023}) enhance dataset quality by focusing generation on failure cases, hard examples, or model weaknesses, often using automated weighting or human-in-the-loop validation to filter noisy or low-quality outputs.
\end{itemize}

Studies also note that the importance of maximizing diversity decreases when synthetic data is mixed with even modest amounts of real data, since the real samples anchor the data distribution and provide natural variety~\cite{yu_regen_2023}. Furthermore, recent works encourage adaptive or hybrid prompting, along with selective down-weighting of suspected noisy synthetic examples during model training~\cite{gao_self-guided_2023}, to further enhance robustness.

Table~\ref{tab:synthetic_data_methods} summarizes several representative prompt-based synthetic data generation methods, along with their target tasks and main contributions.

\begin{longtable}{|p{3cm}|p{2cm}|p{10 cm}|}
\caption{Representative LLM-based synthetic text data generation methods and their core contributions. This table reveals the rapid evolution of strategies—from prompt engineering and task bootstrapping to diverse augmentation and self-instruction—demonstrating how LLMs now support cost-effective, scalable data creation across multiple NLP tasks. It also highlights common themes such as mixing synthetic and real data, leveraging iterative refinement, and targeting low-resource or complex tasks to maximize model improvements.}
\label{tab:synthetic_data_methods}\\

\hline
\textbf{Approach / Study} & \textbf{Task / Domain} & \textbf{Key Idea and Contribution}\\
\hline
\endfirsthead

\hline
\multicolumn{3}{|c|}{\textbf{Continuation of Table \ref{tab:synthetic_data_methods}}}\\
\hline
\textbf{Approach / Study} & \textbf{Task / Domain} & \textbf{Key Idea and Contribution}\\
\hline
\endhead

\hline
\endfoot

\hline
\multicolumn{3}{|c|}{\textbf{End of Table}}\\
\hline\hline
\endlastfoot

\textbf{A Survey of Data Augmentation} \citep{wang_survey_2024}
  & \textit{General NLP} (various tasks)
  & Comprehensive survey of traditional and neural text augmentation methods, laying groundwork for modern LLM-based augmentation. Highlights early techniques (synonym replacement, back-translation) and their impact.\\
\hline

\textbf{Is GPT-3 a Good Data Annotator?} \citep{ding_is_2023}
  & \textit{Text Classification} (SST-2, etc.)
  & Empirical study on using GPT-3 to \textbf{annotate data}. Found GPT-3 can generate labeled data cheaply and quickly, yielding reasonably high model accuracy, though not fully matching human data quality. Suggests LLM-generated labels are a cost-effective alternative to manual annotation.\\
\hline

\textbf{GPT3Mix} \citep{yoo_gpt3mix_2021}
  & \textit{Text Classification} (low-resource)
  & Proposed mixing GPT-3 generated examples with real data to improve low-resource classification. Uses prompt-based generation to create new examples that are then \textbf{mixed} into training. Demonstrated improved accuracy on several classification benchmarks with limited real data.\\
\hline

\textbf{WANLI} \citep{liu_wanli_2022}
  & \textit{Natural Language Inference}
  & Created a \textbf{synthetic NLI dataset} without human labeling. Used an LLM to generate challenging NLI examples (with label inference) focusing on corners of the feature space. Models trained on WANLI plus standard data showed improved generalization to tougher NLI examples.\\
\hline

\textbf{InPars} \citep{bonifacio_inpars_2022}
  & \textit{Information Retrieval (QA)}
  & Used a T5-based LLM to generate \textbf{search queries} from passages (inverse of typical QA). These synthetic query–passage pairs augmented the training of retriever models, significantly boosting performance on web search and QA retrieval tasks. Demonstrates using LLMs to generate realistic information-seeking queries.\\
\hline

\textbf{Unnatural Instructions} \citep{honovich_unnatural_2023}
  & \textit{Instruction-Following} (general NLP)
  & Produced a large dataset of \textbf{synthetic instruction-output pairs} by prompting GPT-3 with task instructions (not from users, hence "unnatural"). Aimed to supplement scarce human-written instructions for training instruction-following models. This work was foundational for models like \textbf{Alpaca} and similar contemporaries.\\
\hline

\textbf{Self-Instruct} \citep{wang_self-instruct_2023}
  & \textit{Instruction \& Multi-task data}
  & A bootstrapping algorithm where a pretrained LM generates new tasks and solutions from a few seed examples. Initially applied to create \textbf{Stanford Alpaca} dataset (52k instructions) using text-davinci-003. This method showed that an LLM can recursively teach itself new tasks, yielding high-quality multi-turn instruction data without manual writing. Inspired many subsequent works in synthetic instruction generation for both text and code.\\
\hline

\textbf{Empirical Study} \citep{li_synthetic_2023}
  & \textit{Text Classification} (6 tasks)
  & Systematic analysis of GPT-3.5 generated data for classification. Showed \textbf{significant gains (3–26\%)} with synthetic augmentation at 100 real samples, diminishing returns as data increases. Compared prompting methods: finding that \textbf{topic-guided generation} improves diversity and performance. Provided practical recommendations on how much data to generate and which prompting strategy to use in various scenarios.\\
\hline

\textbf{AugGPT: Leveraging ChatGPT for Text Data Augmentation } \citep{dai_auggpt_2023}
  & \textit{Few-shot text classification (Amazon Reviews, PubMed20K, Symptoms Dataset)}
  & AugGPT rephrases sentences into multiple diverse yet semantically consistent versions, improving training data quality.
    Evaluation shows AugGPT significantly improves model accuracy compared to traditional augmentation methods (e.g., synonym replacement, back-translation). Findings suggest ChatGPT-generated synthetic data is a scalable, high-quality augmentation strategy, making it a cost-effective alternative to manual data expansion in NLP.\\
\hline

\textbf{Self-Ask} \citep{press_measuring_2023}
  & \textit{Multi-hop Question Answering (QA), Reasoning}
  & Introduced an LLM-based self-asking strategy where models recursively break down complex questions into simpler sub-questions. Demonstrated improved reasoning performance on multi-step inference tasks by leveraging self-generated sub-questions for enhanced accuracy.\\
\hline

\textbf{CoT} \citep{fan_chain--thought_2023}
  & \textit{Scientific QA, Logical Reasoning}
  & Examined how Chain-of-Thought (CoT) prompting improves LLM reasoning capabilities. Proposed Sci-CoT, a two-stage framework where intermediate rationales are generated separately from answer inference, improving efficiency and model performance.\\
\hline

\end{longtable}

\subsection{Quality and properties of generated data}

While prompt-based generation can yield large quantities of data, ensuring the \textbf{quality and usefulness} of that data is crucial. A major concern is that LLMs may produce \textit{incorrect or nonsensical outputs}, especially in knowledge-intensive tasks or if asked to generate beyond their training distribution. Two common strategies to improve quality are: (1) integrating external knowledge or constraints into generation, and (2) post-generation filtering or selection.

\begin{itemize}
    \item Retrieval-Augmented Generation - 
    One way to ground LLM outputs in reality is to provide relevant facts from a knowledge base or corpus during the prompt. This has given rise to \textit{retrieval-augmented data augmentation} \cite{chai_text_2025}. For example, for a question-answer pair generation, instead of having the LLM make up an answer which might be incorrect, one can retrieve a passage from Wikipedia and prompt the LLM to generate a question whose answer is contained in that passage (and maybe also generate the answer). This way, the synthetic QA pair is guaranteed to be grounded in a real passage, improving factual correctness. Studies have shown that introducing such external knowledge significantly enhances the \textbf{faithfulness} of generated data. The LLM essentially uses the retrieval as additional context, reducing hallucination. Retrieval-based methods have been used for tasks like question answering, dialogue (retrieving profiles or factual info to ground responses), and even for generating fact-rich classification data.
    
    \item Post-Processing and Filtering - 
    After generation, it is common to apply filters or selection criteria to remove low-quality synthetic data. Simple heuristics include removing exact duplicates, checking for prompt leakage (e.g., the LLM copying the input example in a few-shot prompt verbatim), or enforcing output format correctness \cite{yu_large_2023}
    . More advanced techniques train a \textbf{critic or classifier} to judge the quality of synthetic examples
    \cite{ding_is_2023}
    . For instance, one could train a classifier to distinguish real vs. synthetic data and then drop synthetic examples that are too easily identified as fake (assuming those are less realistic). Another approach is to only keep synthetic examples on which a strong model (or the LLM itself) predicts the label with high confidence, on the intuition that obvious or easy synthetic examples are likely correct.     
    Ding et al. asked whether GPT-3 is a good annotator \cite{ding_is_2023}; part of answering that is identifying when GPT-3's generated label or text is likely wrong. They and others have used \textit{agreement among multiple model outputs} or consistency checks as signals: if an LLM produces contradictory answers when rephrasing the query, the data point might be unreliable.     
    A notable approach to boost quality is to incorporate \textbf{human feedback or evaluation in the loop}. While the goal is to avoid heavy human labor, having humans quickly review a sample of synthetic data can provide an estimate of quality and help calibrate automated filters. Some studies employed crowdworkers to rate the correctness of synthetic QA pairs or the grammaticality of synthetic sentences, and then used those ratings to train automatic filters (e.g., a model to predict the score) \cite{maufe_pipeline_2022}. This semi-automatic process aims to capture human judgment at scale.
\end{itemize}

\textbf{Distribution Alignment} 
Another subtle issue is aligning the synthetic data distribution with the real data distribution of the task. Synthetic data that is too \textit{out-of-distribution} may not actually help the model perform better on real data. For example, an LLM might generate very pedantic or overly complex sentences that, while correct, do not resemble the style of text the model will see at test time. Techniques like the \textit{Step-by-Step} method mentioned earlier explicitly target alignment by focusing on misclassified real examples \cite{shinn_reflexion_2023}. Additionally, some works (e.g., \cite{zhang_effective_2023}  in the context of topic-guided generation) aim to ensure the synthetic data covers the same topical space as the real data. If real data topics are unknown (in zero-shot cases), one might use broad coverage generation and then filter to those that seem plausible.

\textbf{Synthetic QA and Reasoning} 
An emerging area is using LLMs to generate not just answers but also \textbf{explanations or reasoning chains} for QA tasks. For instance, to train models that perform multi-hop reasoning, one can prompt an LLM to generate a question, the correct answer, and a step-by-step explanation (chain-of-thought \footnote{Chain-of-thought (CoT) prompting guides models to produce intermediate reasoning steps or explanations before providing a final answer~\cite{fan_chain--thought_2023}.}). This produces synthetic training data for both the task and the format of reasoning.
However, ensuring the correctness of the chain-of-thought is challenging; often an LLM might reach a correct answer via a flawed explanation, or vice versa. Therefore, researchers sometimes constrain generation by verifying final answers or using tools (like calculators, knowledge bases) in the loop to confirm intermediate facts.

\subsection{Final remarks}

In summary, synthetic text data generation with LLMs is a \textbf{balancing act between quantity and quality}. With only prompting, it is easy to get a lot of data quickly, but careful design is needed to ensure that data is diverse, relevant, and correct. Augmenting generation with retrieval, iteratively refining prompts based on model errors, and filtering outputs are all effective ways to improve the final dataset. Empirical evidence suggests that these measures can make synthetic data \textit{nearly as effective as real data} in some scenarios, especially for augmenting small human-curated datasets.

We now turn to the code domain, where similar ideas are being applied, with some unique twists due to the nature of code.

\section{Synthetic Data Generation for Code Tasks}

Generative models for programming code – often referred to as \textbf{Code LLMs} when specialized – have also benefited from synthetic data generation \cite{yang_intercode_2023, piterbarg_training_2025}
. The code domain introduces distinct challenges and opportunities. Unlike free-form text, code has a rigid syntax and an execution semantics, allowing for automatic verification of correctness (a property rarely available for natural language) \cite{yu_what_2024}. This section reviews how LLMs are leveraged to generate synthetic code data for training models on tasks such as code completion, code translation, bug fixing, and coding QA \cite{le_coderl_2022, jain_llm-assisted_2023}. We highlight methods that exploit execution feedback \cite{haluptzok_language_2023}, the creation of large synthetic coding instruction datasets \cite{cassano_knowledge_2024}, and strategies to handle language-specific or low-resource coding scenarios \cite{liu_survey_2024}.

\subsection{Various approaches for Prompt-Based Synthetic Code Generation}

Large language models capable of coding (e.g., OpenAI Codex, CodeGen, Code-Llama, StarCoder) are typically trained on massive corpora of existing code from repositories. However, researchers have started to \textbf{supplement real code corpora with synthetic code} generated by LLMs to address specific gaps or to fine-tune models on particular tasks \cite{chen_mastering_2025}. Synthetic code data generation can take multiple forms, as creation of instruction-based datasets, code translation and refactoring, generating problem specifications and solutions.

\subsubsection{Instruction-Following approaches}
\label{sec:intrctionFollowing}

One major use of LLMs is to create QA or instruction-based datasets for code. For example, \textbf{Code Alpaca} is a synthetic dataset of 20K coding instruction-following examples (instruction and solution) generated by applying the Self-Instruct method to code tasks \cite{chaudhary_sahil280114codealpaca_2025}. Starting from a handful of seed prompts (like “Write a Python function to ...”), an LLM (ChatGPT) was used to generate new programming instructions and their solutions, across various domains (file I/O, algorithms, web development, etc.). This dataset was used to fine-tune smaller models (similar to how the Alpaca dataset was used for text \cite{chaudhary_sahil280114codealpaca_2025, jiang_survey_2024} 
), producing code assistants without needing OpenAI’s proprietary data.

\textbf{WizardCoder} \cite{luo_wizardcoder_2023} extended this idea by using an \textit{evolutionary strategy} to increase instruction complexity. They iteratively prompted ChatGPT to produce slightly more complex variations of coding tasks and solutions (Code Evol-Instruct), aiming to enrich the diversity and difficulty of the synthetic data beyond what straightforward prompting would yield.

Another, \textbf{Magicoder} \cite{wei_magicoder_2024}, focused on open-source code: it collected code snippets from GitHub and then generated diverse instruction prompts that would lead to those code snippets as answers, yielding 75K synthetic instruction-output pairs. 
The dataset was generated using ChatGPT as the LLM, which was prompted to create programming instructions corresponding to the collected code snippets. All these efforts show a trend of using LLMs to generate \textit{specialized fine-tuning data for code LLMs}, analogous to how synthetic instruction data is used to fine-tune general LLMs for better following user requests.

\subsubsection{Code Translation and Refactoring}
\label{sec:codeTranslation}

LLMs can be used to generate parallel code pairs, for example translating code from one programming language to another. \textit{Synthetic parallel data} is valuable in training code translation models (transpilers) or enhancing multilingual code understanding. A study by Lachaux et al.
\cite{lachaux_unsupervised_2020}
on TransCoder had initially used unsupervised translation; now with LLMs, one can prompt a model like GPT-4 to translate a given code snippet from Python to Java, etc., creating high-quality pairs.

Similarly, LLMs can perform \textbf{code refactoring}: generating a cleaned-up or optimized version of a given code. \cite{jain_llm-assisted_2023} introduced a pipeline to improve code dataset quality by systematically refactoring code – renaming variables, formatting, adding comments – to make multiple versions of the same program that are functionally identical. 
These transformed versions act as synthetic data that can help a model learn to generalize across different coding styles and improve readability.
The CodeLLaMa-7B model was used for these transformations, as the study found that fine-tuning CodeLLaMa-7B on the cleaned-up versions of programs improved performance by up to 30\% compared to fine-tuning on unprocessed datasets. 

Another approach, \textbf{LintSeq} \cite{piterbarg_training_2025}, generates synthetic data by taking real code and creating a sequence of incremental edits (like commit changes) to reach a goal. The CodeLLaMa-7B model was used to generate these edits, systematically modifying existing code while preserving its functionality.
Models trained on these edit sequences learned to produce more \textbf{diverse solutions} when asked to generate code, as they implicitly saw many ways to modify and improve code \cite{chen_mastering_2025}. Such data is particularly useful for code generation tasks where there are multiple valid solutions (e.g., competitive programming can have many different correct programs for the same problem).

\subsubsection{Problem Generation and Solution Synthesis}
\label{sec:problemGeneration}

Another line of work uses LLMs to generate entire \textbf{programming problems} (specifications) and their solutions. This is akin to how one might generate QA pairs for text; here we generate challenge prompts for coding. For instance, an LLM might be asked: \textit{“Create a coding challenge about graph traversal”} and then also asked to produce a reference solution code. These synthetic challenges can augment datasets like competitive programming problems, which are otherwise limited by human-curated problem sets.

Microsoft’s \textbf{AlphaCode} \cite{li_competition-level_2022}
and DeepMind’s \textbf{AlphaDev} \cite{mankowitz_faster_2023}
efforts, while primarily focused on solving problems, hint at generating many candidate solutions and sometimes even new problems to test on.AlphaCode was developed using a fine-tuned version of CodeGen \cite{nijkamp_codegen_2023}, while AlphaDev leveraged reinforcement learning with AlphaZero-inspired techniques to discover more efficient sorting algorithms.  

A related concept is generating \textbf{buggy code} and fixes: by intentionally introducing bugs into correct code using an LLM (or vice versa), one can create training data for bug detection and repair systems. In a recent preprint, \textit{“LLM-itation is the Sincerest Form of Data”}  \cite{leinonen_llm-itation_2024}, researchers generated synthetic buggy student code submissions for programming education by prompting GPT-4 to imitate typical mistakes students make \cite{leinonen_llm-itation_2024}. They found that the distribution of errors (test case failures) in the synthetic buggy code was statistically similar to real student code, indicating that LLMs can realistically model the space of \textbf{incorrect solutions}. This synthetic data can help create better automated tutors and grading systems without needing to collect tons of real student submissions (which are privacy-sensitive).

Similarly, for automated code repair, one can generate (buggy code $\rightarrow$ fixed code) pairs: \cite{haluptzok_language_2023} proposed a self-improvement strategy where a model 
generates its own code output, then attempts to identify errors (perhaps via unit tests or linting) and repair them, producing (initial, fixed) code pairs for training another model \cite{liu_best_2024}. The approach utilized GPT-3 as the base LLM for generating buggy and corrected code samples, which were verified using a Python interpreter before being incorporated into training data. 

Table \ref{tab:synthetic_code_data} provides examples of LLM-driven synthetic data initiatives in the code domain and their focus.

\begin{longtable}{|p{3.2cm}|p{4.2cm}|p{8.6cm}|}
\caption{Overview of representative LLM-based approaches for synthetic code data generation, covering tasks such as code generation, translation, refactoring, instruction tuning, and bug synthesis. \newline
\textbf{Note:} Where “No specific benchmark” is listed, the method was evaluated on custom or task-specific datasets rather than a standard public benchmark.}
\label{tab:synthetic_code_data}\\

\hline
\textbf{Approach / System} & \textbf{Code Task Domain, Input Type} & \textbf{Key Idea, Contribution, and Benchmark/Dataset} \\
\hline
\endfirsthead

\hline
\multicolumn{3}{|c|}{\textbf{Continuation of Table \ref{tab:synthetic_code_data}}} \\
\hline
\textbf{Approach / System} & \textbf{Code Task Domain, Input Type} & \textbf{Key Idea, Contribution, and Benchmark/Dataset} \\
\hline
\endhead

\hline
\endfoot

\hline
\multicolumn{3}{|c|}{\textbf{End of Table}} \\
\hline\hline
\endlastfoot

\textbf{CodeRL} \citep{le_coderl_2022}
  & Code generation with RL (\textit{Generation of problem specification and solution}); Input: Code2Code
  & Uses actor-critic RL on code generation. The LLM generates code; execution feedback (does the code run?) is used as a reward to refine the model. \textbf{Benchmarks:} HumanEval\citep{chen_evaluating_2021}, MultiPL-E\citep{hendrycks_measuring_2021} \\
\hline

\textbf{Self-Code-Improve} \citep{haluptzok_language_2023}
  & Code translation, refactoring, optimization, repair; Input: Code2Code
  & LLM generates code and its chain-of-thought, then analyzes for errors or inefficiencies (simulated self-debugging with heuristic prompts). \textbf{Benchmark:} No specific benchmark; aligns with execution-based debugging tasks. \\
\hline

\textbf{InterCode} \citep{yang_intercode_2023}
  & Interactive code generation with RL (\textit{specification + solution}); Input: Code2Code
  & Combines LLMs with an interactive environment: code as actions, execution feedback as observations. \textbf{Benchmark:} No specific benchmark; execution-based evaluation. \\
\hline

\textbf{Reflexion} \citep{shinn_reflexion_2023}
  & Self-refinement in code generation; Input: Code2Code
  & LLM outputs an initial solution, then refines it based on feedback, creating (initial, refined) pairs. \textbf{Benchmark:} HumanEval\citep{chen_evaluating_2021} \\
\hline

\textbf{Code Alpaca} \citep{chaudhary_sahil280114codealpaca_2025}
  & Code instruction following (\textit{instruction dataset creation}); Input: Text2Code
  & 20K dataset of synthetic coding instructions and answers via Self-Instruct with ChatGPT. \textbf{Benchmark:} No specific benchmark; instruction-based evaluations. \\
\hline

\textbf{WizardCoder} \citep{luo_wizardcoder_2023}
  & Complex code instructions (\textit{instruction dataset creation}); Input: Text2Code
  & Builds on Code Alpaca, using evolutionary approaches for more complex, diverse instructions. \textbf{Benchmarks:} HumanEval\citep{chen_evaluating_2021}, CoderEval\citep{yu_codereval_2024} \\
\hline

\textbf{Magicoder – OSS-Instruct} \citep{wei_magicoder_2024}
  & Code instruction tuning; Input: Text2Code
  & Created 75K synthetic Q\&A pairs from open-source code. \textbf{Benchmark:} No specific benchmark; aligns with instruction-tuned datasets. \\
\hline

\textbf{Test-Translating for Low-Resource} \citep{cassano_knowledge_2024}
  & Multilingual code generation, translation; Input: Code2Code
  & Generates code for low-resource languages by translating and verifying high-resource code. \textbf{Benchmark:} No specific benchmark; multilingual evaluations. \\
\hline

\textbf{NL2SQL Augmentation} \citep{liu_survey_2024}
  & Text-to-SQL parsing (\textit{instruction dataset creation}); Input: Text2Code
  & Combines strong and weak LLMs to generate diverse SQL queries. \textbf{Benchmark:} No specific benchmark; text-to-SQL evaluation. \\
\hline

\textbf{PanGu-Coder} \citep{christopoulou_pangu-coder_2022}
  & Code generation (\textit{instruction dataset creation}); Input: Text2Code
  & Large-scale LLM for code generation (Huawei). \textbf{Benchmark:} No specific benchmark. \\
\hline

\textbf{CodeGen} \citep{nijkamp_codegen_2023}
  & Code generation and completion (\textit{instruction dataset creation}); Input: Text2Code
  & Open-source model for text-to-code generation in multiple languages. \textbf{Benchmarks:} HumanEval\citep{chen_evaluating_2021}, MultiPL-E\citep{hendrycks_measuring_2021} \\
\hline

\textbf{InCoder} \citep{fried_incoder_2023}
  & Code completion and synthesis (\textit{instruction dataset creation}); Input: Text2Code
  & Generative model for autoregressive code snippet completion. \textbf{Benchmarks:} HumanEval\citep{chen_evaluating_2021}, DS-1000\citep{lai_ds-1000_2023} \\
\hline

\textbf{AlphaCode} \citep{li_competition-level_2022}
  & Competitive programming (\textit{problem spec \& solution}); Input: Text2Code
  & Generates multiple solutions, ranks by correctness (Microsoft). \textbf{Benchmarks:} HumanEval\citep{chen_evaluating_2021}, CoderEval\citep{yu_codereval_2024} \\
\hline

\textbf{Codex} \citep{chen_evaluating_2021}
  & General code generation (\textit{instruction dataset creation}); Input: Text2Code
  & OpenAI's Codex (GitHub Copilot), pretrained on both language and code. \textbf{Benchmarks:} HumanEval\citep{chen_evaluating_2021}, MultiPL-E\citep{hendrycks_measuring_2021} \\
\hline

\textbf{TransCoder} \citep{lachaux_unsupervised_2020}
  & Code translation (\textit{code2code}); Input: Code2Code
  & Unsupervised machine translation for programming languages. \textbf{Benchmark:} No specific benchmark; multilingual evaluation. \\
\hline

\textbf{LintSeq} \citep{piterbarg_training_2025}
  & Incremental code modifications (\textit{code2code}); Input: Code2Code
  & Generates synthetic data by creating incremental edits of real code. \textbf{Benchmark:} No specific benchmark; edit-based evaluation. \\
\hline

\textbf{AlphaDev} \citep{mankowitz_faster_2023}
  & Optimized code synthesis (\textit{problem spec \& solution}); Input: Code2Code
  & RL-based model for efficient sorting algorithms (DeepMind). \textbf{Benchmark:} AiXBench\citep{hao_aixbench_2022} \\
\hline

\textbf{LLM-itation (Buggy Code Generation)} \citep{leinonen_llm-itation_2024}
  & Buggy code and fix generation (\textit{instruction dataset creation}); Input: Text2Code
  & Uses LLMs (GPT-4) to synthesize buggy student code for tutoring. \textbf{Benchmark:} No specific benchmark; educational and bug-detection datasets. \\
\hline

\end{longtable}

\subsection{Quality and properties of generated data}

\textbf{Execution Feedback and Validation} A key advantage in the code domain is that the correctness of generated data can often be \textit{automatically checked}. This is typically done by running the code (if possible) or using tests and compilers. Execution feedback serves multiple roles:

\begin{itemize}
    \item \textbf{Filter Correct Solutions:} When generating code solutions, one can run each candidate on a set of unit tests. Only those that pass all tests are kept as synthetic data. This ensures the synthetic set consists of functionally correct programs. As a result, a model trained on this data learns from mostly correct examples, which is important because learning from incorrect outputs could mislead it. Some projects use an LLM to propose dozens of solutions to a coding task and then filter by execution to build a high-quality dataset of solutions.

    \item \textbf{Learn from Failure:} Conversely, the patterns in \textit{failed executions} can also be informative. As seen in Reflexion and related methods
    \cite{shinn_reflexion_2023}, if a code fails, the error messages or partial progress can be fed back into the generation loop. This can yield a synthetic sequence: (attempt1, error, attempt2 fixed, error2, ..., final correct). Training on such sequences can teach a model to perform iterative refinement. Moreover, analyzing where the LLM's outputs tend to fail (e.g., always off-by-one in a loop, or common syntax mistakes) highlights where model logic is weak. One can then prompt the LLM to specifically generate examples around those tricky cases, turning model failure modes into new training data to correct those modes.

    \item \textbf{Reward Modeling:} Execution results provide a natural reward signal for reinforcement learning. CodeRL \cite{le_coderl_2022}, for instance, uses whether the code executes correctly as a reward to fine-tune the model via RL (reinforcement learning). Instead of supervised training on static data, the model \textit{explores} code solutions and is reinforced to produce outputs that pass tests. Over time, the distribution of generated code shifts towards higher correctness. This approach effectively generates and uses synthetic data on the fly in a closed loop.

    \item \textbf{Measuring Difficulty:} By using execution feedback, one can also stratify the synthetic data by difficulty. Solutions that required multiple attempts or where most LLMs failed except one might indicate a hard problem. These can be labeled as such and a model can be trained to handle varying difficulty by exposing it to a mix (or by curriculum learning from easier synthetic tasks to harder ones).
\end{itemize}

Using execution signals is somewhat analogous to \textit{fact-checking} in text generation – just as a fact-based QA might check an answer against a source text, a code generator checks a program against test cases. The difference is the test oracle for code is often definitive (pass/fail), whereas in language tasks, factual correctness can be fuzzy. This makes synthetic code generation uniquely powerful: one can generate vast amounts of code and reliably keep only the correct and relevant ones, yielding a high-precision training set \cite{liu_best_2024}.

\textbf{Addressing Biases and Data Coverage in Code} 
Synthetic code generation also needs to consider diversity and bias, albeit in a different sense than text. \textbf{Bias in code data} might refer to over-representation of certain patterns (e.g., always using a certain library) or style (e.g., code that always uses list comprehensions versus loops). LLMs might learn these biases from training data and also reflect them in synthetic outputs. To counter this, some strategies include:

\begin{itemize}
    \item \textbf{Generating the same functionality in multiple ways} (as the refactoring approaches do) to prevent the model from thinking there's only a single correct coding style. For instance, generate both iterative and recursive solutions for a problem if both are valid.
    
    \item \textbf{Ensuring comment diversity} and variable naming diversity in synthetic code, so the model doesn't overly rely on specific token sequences that might correlate with certain outputs.

    \item \textbf{Covering edge cases:} If real code data is mostly clean, an LLM might not naturally produce code with unusual constructs. But for robustness, one might want synthetic examples that include, say, very deeply nested loops, or unusual combinations of language features. This can be prompted explicitly (e.g., ``Write a program to X that uses recursion and bit manipulation'').

    \item \textbf{Bias towards older or newer practices:} Code corpora often contain outdated patterns. An LLM might produce deprecated syntax if trained on old data. Synthetic generation can intentionally produce updated code (for example, use Python f-strings instead of older \% formatting) by instructing it accordingly, thus helping models stay modern. Conversely, it could also simulate legacy code if needed for training a model to understand older code.
\end{itemize}

In terms of data coverage, one challenge is the long tail of programming tasks. Real code datasets might have many 
examples of common tasks (sorting, web requests) but few of niche ones (like bit-level hacks or graphics shaders). 
LLMs can be directed to generate scenarios for these tail tasks, effectively filling gaps. Some surveys like 
\textit{Mastering the Craft of Data Synthesis for CodeLLMs} categorize over 50 recent works across 23 topics in 
code synthetic data \cite{chen_mastering_2025}, indicating a rich taxonomy of methods addressing different areas 
(from data structure manipulations to API usage scenarios). It shows that the community is actively ensuring that 
synthetic data methods cover a wide spectrum of programming needs.

\subsection{Final remarks}

Empirical evidence suggests that synthetic code data, when used appropriately, can significantly improve code-focused models. Several open-source code LLMs (e.g., variations of Code-Llama, StarCoder) have integrated synthetic instruction tuning data and reported better benchmark results in code generation and completion tasks after fine-tuning on such data. One analysis noted that models which used primarily real code for pre-training but \textbf{synthetic data for fine-tuning (especially for instruction following)} often outperform those fine-tuned on limited human data, due to the \textbf{abundance and diversity} of the synthetic data. For instance, Qwen-2.5-Coder (hypothetical name from context) favored a pipeline where the base model is trained on GitHub code and then \textbf{instruction-tuned on a large synthetic set}, finding this more efficient to achieve alignment than gathering human-written instructions.

In specialized tasks like SQL query generation (NL2SQL \cite{liu_survey_2024}), using LLMs to generate extra training pairs has advanced the state of the art \cite{chen_mastering_2025}. By generating varied NL questions for existing databases and ensuring their correctness (through execution on the database or a SQL validator), researchers created training data that taught models to handle more linguistic variations and database schemas, yielding higher accuracy on text-to-SQL benchmarks.

Overall, synthetic data generation for code is quickly becoming an indispensable part of the toolkit for building code AI systems. The ability to get infinite ``free'' coding examples, combined with the power to verify them, allows for creative training regimes that were not possible with just the raw code found in the wild. As with text, though, caution is needed to ensure synthetic code is \textbf{representative, correct, and useful} for the target tasks.

\section{Challenges and Considerations}

Despite the progress and successes of LLM-based synthetic data generation, several challenges persist. We outline the key issues and research considerations when applying these techniques, applicable to both text and code domains (with some domain-specific nuances where appropriate).

\subsection{Quality Assurance and Factuality}
Ensuring the \textit{factual correctness} (for text) and \textit{functional correctness} (for code) of synthetic data is paramount. LLMs are prone to \textbf{hallucinations} -- producing incorrect facts or invalid code confidently. If not filtered out, these errors can propagate into the models trained on such data, causing them to learn spurious patterns. For text data, hallucinated content might teach a model incorrect knowledge or bias; for code, learning from non-working code could directly degrade a model’s ability to generate runnable programs.

\textit{Mitigations}: As discussed, integrating retrieval to ground text generation \cite{chai_text_2025} and execution to validate code \cite{liu_best_2024} are primary ways to tackle this. Another line of defense is using \textbf{evaluation metrics} or critics. For example, using perplexity under a strong language model as a proxy for quality: if a GPT-4 model finds a GPT-3.5-generated sentence to be very unlikely, that sentence might be of low quality. Similarly, static analyzers or linters can catch obvious issues in code (like syntax errors or insecure practices) to filter those out. Human review of a subset can also estimate the precision of synthetic data; if it’s too low, one might tighten generation criteria. Recent research also develops metrics for \textit{faithfulness} of generated content, which could be employed to score and select synthetic examples.

\subsection{Distribution Shift and Realism}
Synthetic data, no matter how well produced, may not perfectly follow the distribution of real data, especially user-generated data. This mismatch—known as \textit{distribution shift}\footnote{Distribution shift denotes a mismatch between the data distribution of synthetic data and that of real data, which can adversely impact model performance~\cite{liu_best_2024}.}—can result in models that are \textbf{overfit to synthetic patterns}. For instance, an LLM might generate cleaner and more coherent text than the often noisy real-world text (full of typos, slang, etc.), and a classifier trained on such immaculate synthetic text might struggle on messy real text. In code, an LLM might prefer a certain style (e.g., always using \texttt{for} loops) and underrepresent other valid styles, causing a model to be biased.

\textit{Mitigations}: Mixing synthetic data with real data is a common practice to alleviate this \cite{li_data_2024}. Maintaining a core of real examples can \textit{anchor} the model in reality, while the synthetic portion provides augmentation. Some works explicitly try to \textit{match the distributions}: e.g., generate synthetic data that is indistinguishable from real data according to some classifier (an adversarial approach). In the extreme, \textbf{Generative Adversarial Networks (GANs)} style training could be considered, though with text that is tricky to do in practice. Another mitigation is \textbf{diversity enhancement} -- ensuring synthetic data itself is diverse enough so that the model doesn’t get a narrow view. Using multiple LLMs or multiple prompts to generate a variety of styles can help. Also, curriculum strategies (train first on some real, then gradually add synthetic) might prevent the model from skewing too far.

A related concern is \textbf{model bias amplification}. If the LLM used for generation has certain biases (e.g., it might generate more male characters in stories, or prefer certain political slants), those carry into the synthetic data and can even amplify if the data is used to train a smaller model which then is used to generate more data, and so on. Careful prompt design (e.g., instructing the LLM to ensure diverse representation) and post-hoc balancing of the synthetic dataset are needed to counteract this.

\subsection{Evaluation of Synthetic Data and Models}
\label{sec:evaluate-synthetic}

Evaluating synthetic data and the models trained on it is non-trivial but crucial to ensure that synthetic augmentation leads to real, measurable improvements. Both the \textit{quality of the data} and the \textit{performance of models} trained with it must be assessed.

We summarize the principal criteria and strategies as follows:

\begin{itemize}
    \item \textbf{Downstream Task Performance:} The most direct method is to train a model using (or fine-tuned with) synthetic data and measure its performance (accuracy, F1, BLEU, ROUGE, etc.) on a real, human-annotated test set. For code, this means pass@k, HumanEval, and related benchmarks.
    \item \textbf{Ablation/Comparison Experiments:} Controlled experiments with (a) real data only, (b) synthetic only, (c) mixed, to quantify the benefit of synthetic data.
    \item \textbf{Diversity and Realism Metrics:} Distinct-n (n-gram diversity), Self-BLEU, perplexity, and distributional similarity to real data. For code, use execution-based correctness, error analysis, and expert review.
    \item \textbf{Distribution Shift Analysis:} Classifier-based discrimination between real and synthetic samples can expose distribution gaps; manual inspection and error analysis remain invaluable.
    \item \textbf{Human Evaluation:} For generative tasks, human annotators (experts or crowdworkers) can judge correctness, fluency, and usefulness, often using Likert scales or pairwise preferences.
    \item \textbf{Robustness and Generalization:} Test models against out-of-distribution samples or rare/edge cases to reveal potential overfitting to synthetic patterns.
    \item \textbf{Specialized Code Metrics:} For code, also consider benchmarks like HumanEval, CoderEval, MultiPL-E, and DS-1000, as well as metrics like execution accuracy, logical form accuracy (for text-to-SQL), and detailed error types (syntax, logic, etc.).
\end{itemize}

\paragraph{Statistical Rigor and Robustness.}
To meaningfully quantify the robustness and generalizability of improvements attributed to synthetic data, we strongly encourage the adoption of statistical testing and rigorous validation practices in future work. Key approaches include reporting confidence intervals (e.g., via bootstrapping or repeated trials), conducting cross-dataset validation (training and evaluating models on diverse datasets or splits), and employing formal statistical tests (such as paired t-tests or Wilcoxon signed-rank tests) when comparing models trained on synthetic versus real or hybrid data. Such practices not only improve the transparency and reliability of reported gains, but also help identify cases where improvements may not generalize across tasks, domains, or data distributions. We recommend that future benchmarks and empirical studies on synthetic data explicitly include these statistical safeguards as standard methodology.

Recent best practices recommend always providing: (1) clear comparisons with real-data baselines, (2) transparency about synthetic dataset composition, and (3) open reporting of both benefits and risks (e.g., error modes, distribution gaps, overfitting tendencies). Explicitly monitoring for distribution shift and overfitting is especially important in closed-loop synthetic data pipelines.

To assist researchers and practitioners, Table~\ref{tab:synthetic-dataset-benchmarks} provides an overview of widely-used synthetic datasets, associated tasks, and standard evaluation metrics in the literature.

\begin{table*}[ht]
\centering
\caption{Benchmark synthetic datasets, task types, and commonly used evaluation metrics.}
\label{tab:synthetic-dataset-benchmarks}
\begin{tabular}{|l|l|p{3.3cm}|p{4cm}|l|}
\hline
\textbf{Dataset} & \textbf{Domain} & \textbf{Task Type} & \textbf{Evaluation Metrics} & \textbf{Reference} \\
\hline
\textbf{WANLI} & Text & Natural Language Inference (NLI) & Accuracy, F1, Human Eval & \cite{liu_wanli_2022} \\
\textbf{GPT3Mix} & Text & Text Classification (low-resource) & Accuracy, F1, BLEU & \cite{yoo_gpt3mix_2021} \\
\textbf{Unnatural Instructions} & Text & Instruction-following, Multi-task NLP & Model Accuracy, Human Eval & \cite{honovich_unnatural_2023} \\
\textbf{Self-Instruct / Alpaca} & Text & Multi-turn Instruction Following & Model Accuracy, Win Rate, Human Eval & \cite{wang_self-instruct_2023} \\
\textbf{AugGPT} & Text & Text Classification (Augmentation) & Accuracy, F1 & \cite{dai_auggpt_2023} \\
\textbf{Code Alpaca} & Code & Instruction-based Code Generation & Pass@k, HumanEval, CodeEval & \cite{chaudhary_sahil280114codealpaca_2025} \\
\textbf{WizardCoder} & Code & Complex Code Instructions & Pass@k, HumanEval, CoderEval & \cite{luo_wizardcoder_2023} \\
\textbf{AlphaCode} & Code & Competitive Programming & Pass@k, HumanEval, Problem Solved & \cite{li_competition-level_2022} \\
\textbf{Reflexion} & Code & Self-refinement/Repair, Code Generation & Pass@k, HumanEval & \cite{shinn_reflexion_2023} \\
\textbf{NL2SQL Augmentation} & Code/Text & Text-to-SQL Parsing & Execution Accuracy, Logical Form Accuracy & \cite{liu_survey_2024} \\
\textbf{WANLI (Code)} & Code & N/A (mainly text, but sometimes code NLI) & - & - \\
\hline
\end{tabular}
\end{table*}

\subsection{Overfitting and Distribution Shift Risks in Closed-Loop Synthetic Data Generation}
\label{sec:overfitting-risks}

While numerous studies report substantial performance improvements when augmenting training with synthetic data, it is crucial to acknowledge the inherent risks of overfitting and distribution shift—especially in closed-loop generation pipelines, where models are recursively trained on data generated by themselves or by closely related models. In methods such as Self-Instruct for text or AlphaDev for code, the iterative generation and re-training cycle can induce what is known as \emph{model collapse} or self-reinforcement: errors, biases, or artifacts present in the synthetic data may become amplified over successive rounds, leading to degraded generalization or “over-specialization” to synthetic data patterns~\cite{gerstgrasser_is_2024}. For example, empirical studies have shown that models trained exclusively on multi-generation synthetic corpora tend to lose robustness, exhibit distributional drift, or encode spurious correlations not present in real-world data. This effect is most pronounced when real data is scarce or entirely absent from the training mix.

Mitigation strategies include always mixing real and synthetic data (rather than relying solely on synthetic outputs), applying strong filtering and validation to synthetic examples, and adopting curriculum or adversarial training schemes to prevent distribution drift. Recent research demonstrates that as long as a core of real data is maintained and synthetic augmentation is used judiciously, the risk of catastrophic model collapse can be minimized~\cite{gerstgrasser_is_2024, liu_best_2024}. Nonetheless, practitioners should be aware of these pitfalls, conduct careful ablation studies, and explicitly monitor for overfitting and loss of generalization when deploying synthetic data pipelines in practice.

\subsection{Scale and Cost Trade-offs}

While generating data is cheaper than manual labeling, it still incurs cost (especially if using an API like OpenAI’s) and time for large volumes. Deciding \textit{how much synthetic data to generate} is an open question. Generating more might improve coverage but with diminishing returns and increased redundancy \cite{li_data_2024}. In \cite{li_synthetic_2023}, after a certain point, adding more GPT-generated examples didn’t improve the classifier further for some tasks. There is also the question of using more expensive models (like GPT-4) to generate presumably higher-quality data versus using cheaper ones (GPT-3.5) for more quantity. A few works have begun to compare generation quality from different LLMs; early indications are that larger models do give better data (more nuanced, fewer mistakes), but perhaps one can achieve similar results by filtering a larger set from a smaller model. Strategically, one might generate a bulk of data with a fast model and then have a powerful model or humans filter/evaluate it.

From a research perspective, sharing and reproducibility of synthetic data is tricky. If each user just prompts ChatGPT and gets data, it might be slightly different. There are efforts like \textbf{DataDreamer} \cite{patel_datadreamer_2024} that aim to create standardized tools for synthetic data generation with LLMs, so that datasets can be reproduced and experiments repeated consistently. This is essential for scientific rigor.

\subsection{Ethical and Legal Considerations}

Synthetic data, despite being artificial, can still pose ethical issues. If the LLM's training data contained biases, hate speech, or private information, these could surface in synthetic outputs. There have been instances of LLMs leaking bits of memorized training data verbatim. If such data is re-used, it could violate privacy or copyright. Researchers must ensure that no sensitive or copyrighted content is inadvertently \textbf{regenerated} in synthetic datasets \cite{liu_best_2024}. Techniques to detect verbatim regeneration (like Google’s duplex scanning or embedding similarity to known datasets) are used to filter outputs that are too close to any single training example. This is especially important if synthetic data is meant to be openly shared – one must ensure it's not just a regurgitation of proprietary data.

On the flip side, synthetic data can enhance privacy if done right (as noted in motivation), but one should not assume synthetic equals safe by default. For instance, an LLM might produce a very realistic fake patient record. If that record accidentally matches an actual person’s data (due to the model memorizing and outputting it), privacy is breached. The chance is low but non-zero with large models. So, thorough checks or using models fine-tuned to avoid such leakage (some research is happening in this area) are needed.

\textbf{Model collapse} is a more theoretical but increasingly relevant concern: this is the idea that if models are trained on their own outputs repeatedly, the quality degenerates over generations. Think of it as making a copy of a copy of a copy... eventually, the content might become nonsensical. Some studies in 2023-2024 (e.g., \cite{gerstgrasser_is_2024}) have investigated this by simulating multiple rounds of training a model on its predecessor's outputs. They did find that if each new model is trained purely on the previous model's synthetic data (replacing real data entirely), performance drops – confirming the \textit{model collapse} effect \cite{gerstgrasser_is_2024}. However, they also found an encouraging result: if each generation \textit{keeps the real data and just adds some synthetic}, the performance can be maintained or even improved without collapse. In other words, \textbf{accumulating} synthetic data on top of real data avoids the degenerative feedback loop. This suggests that in practice, as long as we don't throw away our original human datasets, we can iteratively augment with model-generated data with less fear of collapse. Nonetheless, vigilance is needed if we foresee a future where a model might be trained on predominantly AI-created corpora (e.g., lots of websites start hosting AI-written content).

\subsection{Task-Specific Nuances}

Different tasks might face unique issues with synthetic data. For example, in dialogue systems, synthetic conversations generated by an LLM might not capture the quirks of human dialogue (like frequent topic changes, misunderstandings, or the richness of multi-turn coherence). Also, user simulators (for task-oriented dialogue) need to behave realistically; an LLM generating both sides of a conversation might become too cooperative, not simulating the true difficulty of dealing with real users. Ensuring that synthetic dialogues are challenging and varied is an ongoing challenge.

In summarization, an LLM can generate an infinite number of summaries for an article, but many will be similar and might not add value beyond what a few reference summaries provide. The challenge is to generate summaries that highlight different possible interpretations or focus points of the article, thereby teaching a summarization model to handle ambiguity or preference.

In code, tasks like generating \textbf{malicious code} (for cybersecurity training) or \textbf{esoteric language code} have their own quirks. An LLM might not know much about a very niche programming language, so generating synthetic data for it might require a different approach (like transpiling known code from a similar language).

\subsection{Human Acceptance}

Finally, a more social challenge: will practitioners trust models trained on synthetic data? There can be a perception that data made by AI is not "real" and thus the model might not work in the real world. Building confidence requires transparency about how the synthetic data was created and evaluated. If synthetic data is to be used in high-stakes areas (medical, legal), one might need regulatory approval or at least proof that using it does not degrade outcomes. Demonstrating equivalence to real data through rigorous tests is important for adoption.

In summary, while synthetic data generation with LLMs is a powerful approach, it must be applied with careful consideration of these challenges. Each challenge is an active area of research, and progress is being made. Next, we look at some of the promising directions for future research that address these challenges and push the boundaries of synthetic data generation.

\section{Future Directions}

The intersection of LLMs and synthetic data is a rapidly evolving field. Based on current trends and open problems, we highlight several future research directions and opportunities:

\begin{itemize}
    \item \textbf{Unified Frameworks and Taxonomies:} As the field grows, there is a need for comprehensive frameworks that unify various techniques. Recent surveys \cite{wang_survey_2024} have started organizing approaches (e.g., by stages of the LLM lifecycle or by data generation vs. curation vs. evaluation), but an agreed-upon taxonomy would aid researchers in identifying gaps. A framework that describes, for instance, the pipeline from prompt design, generation, filtering, to integration into model training could help systematically compare methods and build upon each other’s insights. Projects like \textit{On LLMs-Driven Synthetic Data Generation, Curation, and Evaluation: A Survey} \cite{long_llms-driven_2024} propose generic workflows  – these should be expanded and possibly standardized.

    \item \textbf{Automated Prompt Engineering:} The quality of synthetic data heavily depends on prompt quality. We may see more research into \textbf{automated prompt optimization} for data generation. Techniques like prompt tuning (learnable prompts) or using evolutionary algorithms to evolve prompts that yield better synthetic data could reduce manual trial-and-error. For example, a future system might start with a basic prompt and then automatically adjust it (or try many variants) to maximize downstream validation performance of a model trained on the generated data – essentially tuning the prompt for optimal data utility.

    \item \textbf{Active Learning with LLM Generators:} Integrate LLM data generation into an active learning loop. A model being trained could identify which additional data points would most help it (areas of uncertainty) and request an LLM to generate examples in those areas, possibly with some human vetting. This combines the best of active learning (selecting informative examples) and synthetic generation (actually producing them on demand). Some initial work in this direction might leverage uncertainty sampling to drive prompt topics.

    \item \textbf{Cross-Modal and Multimodal Synthetic Data:} While this survey has focused primarily on text and code, an important direction for future research is the extension of LLM-driven synthetic data generation to multimodal settings—including image-text, audio-text, and even video-text data. Recent advances in vision-language models (e.g., GPT-4V, Gemini, Kosmos) and diffusion-based generative models have made it feasible to synthesize paired data (such as image-caption pairs, synthetic dialogue with associated video frames, or multimodal QA datasets). Synthetic multimodal data can be used to pretrain and fine-tune models for cross-modal retrieval, captioning, audio transcription, and other tasks where real data is scarce or expensive to annotate. Notably, language models can now be prompted to describe non-existent images, generating synthetic captions for use with diffusion models or image generators—effectively creating new datasets for vision-and-language tasks. Similarly, LLMs can generate synthetic speech transcripts or audio scene descriptions, which, when paired with synthesized or real audio, can support research in speech recognition and audio-visual grounding. Despite rapid progress, challenges remain, such as ensuring alignment between modalities, mitigating hallucinated associations, and evaluating the quality of synthetic multimodal data. We expect the coming years to see an expansion of these approaches, with new benchmarks and evaluation criteria specifically tailored to the unique challenges of multimodal synthetic data.

    \item \textbf{Enhancing Real Data with Synthetic Variants:} Instead of generating entirely new data, an alternative is to generate \textit{variants of existing real data}. For example, given a sentence, generate a paraphrase that changes some attributes (sentiment, tense, etc.). This can be more controlled than free generation. Future research might focus on \textit{attribute-conditioned generation}: one explicitly sets what aspects to vary and what to keep. This can create counterfactual datasets for causal analysis or fairness testing (e.g., change demographic references in a sentence to test model bias).

    \item \textbf{Theoretical Understanding:} More theoretical work is needed to understand \textit{why and when} synthetic data helps or hurts. Some initial theoretical analysis addresses model collapse dynamics \cite{gerstgrasser_is_2024} and error accumulation. But a theory for something like "How does the synthetic data distribution interact with the learning algorithm’s generalization error?" would be valuable. Perhaps PAC-learning or domain adaptation theory could be applied: synthetic data can be seen as coming from a distribution that approximates the real distribution; theories of domain shift or distribution blending might provide guarantees or insights. Understanding the limits of augmentation (diminishing returns) in a mathematical sense could guide practitioners on how much is enough.

    \item \textbf{Quality Evaluation Metrics:} Developing better automatic metrics for synthetic data quality will be important. Metrics that correlate strongly with downstream performance would allow tuning generation without having to do full training runs. For example, a metric that combines diversity and fidelity (maybe using pre-trained model embeddings to judge if a synthetic set covers the real set’s embedding space well) could be useful. In code, metrics beyond just "pass rate" might consider code elegance or complexity, depending on what we want the model to learn.

    \item \textbf{Human-in-the-Loop and Expert Knowledge:} Another future direction is incorporating expert rules or human insights directly into generation. An expert could specify domain constraints that the LLM must adhere to when generating data (kind of like guardrails). Or interactive generation where a human quickly corrects or annotates LLM outputs and those corrections feed back into producing better data (perhaps using reinforcement learning from human feedback, RLHF, but applied to data generation policy).

    \item \textbf{Safety Filtering and Bias Mitigation:} Building on ethical concerns, research will likely produce more sophisticated \textit{safety filters} specifically for synthetic data. For instance, if generating a dataset of dialogues, ensuring none of the dialogues turn toxic inadvertently. We might see tools that can enforce certain bias mitigations – e.g., a constraint solver that ensures the synthetic dataset has equal representation of male/female or different ethnic names in certain contexts, to prevent training data bias. LLMs themselves can be asked to \textit{self-debias} outputs (there is some work on prompting models to be mindful of bias). Future synthetic data generators could come with a suite of such self-checks or adjustments.

    \item \textbf{Domain-Specific Synthetic Data Generators:} While generic LLMs are used now, we might have specialized generators fine-tuned specifically for data generation in a domain. For example, a medical note generator LLM (trained on public medical text) that is particularly good at producing plausible medical records for research, or a legal text generator that can produce contracts or case descriptions for legal NLP. These domain-specific LLMs could produce more realistic synthetic data for those fields than a general model like GPT-4. Investing in such specialized synthetic data models might become a trend for companies that need data in domains where real data is hardest to get.

    \item \textbf{Benchmarking and Competitions:} We anticipate the rise of benchmarks specifically for synthetic data generation methods. For instance, a challenge might provide a small real dataset and require participants to produce a synthetic dataset (via an LLM or other methods) that best improves a model's performance on a hidden test set. This would directly measure how effective different generation strategies are. Similarly, competitions for code might provide some base code and encourage generation of the best augmentation set. These would spur innovation and provide comparative evaluations of techniques.

    In essence, the future of LLM-based synthetic data generation is likely to involve more \textbf{automation, integration, and domain-awareness}. As LLMs themselves continue to improve (becoming more factual, less biased, more controllable), the quality of synthetic data we can get from them will also improve. This symbiotic relationship—better models $\rightarrow$ better data $\rightarrow$ even better models—could accelerate AI development. But careful research is needed to ensure we avoid pitfalls (like models reinforcing their own errors). By addressing current challenges and exploring these future directions, we can unlock the full potential of synthetic data in both NLP and code-related fields.
\end{itemize}

To synthesize these emerging research priorities and help guide future work, Table~\ref{tab:future-directions} presents a conceptual mind map of the major directions discussed above. This visual summary highlights key themes—including evaluation, ethics, prompting, multimodality, human-in-the-loop learning, domain-specific generators, and benchmarking—alongside representative subtopics. Such an overview offers readers a quick reference for the multifaceted challenges and opportunities at the forefront of LLM-based synthetic data research.

\begin{table*}[!htbp]
\centering
\caption{Key future directions in LLM-based synthetic data research, grouped by central theme and representative subtopics.}
\label{tab:future-directions}
\begin{tabular}{|l|p{10cm}|}
\hline
\textbf{Theme} & \textbf{Representative Subtopics} \\
\hline
Evaluation            & Robust Metrics, Human Assessment, Distribution Shift \\
Ethics                & Privacy, Bias \& Fairness, Data Leakage \\
Prompting             & Automated Prompt Design, Prompt Tuning, Topic Control \\
Multimodality         & Text-Image, Text-Code, Audio Integration \\
Human-in-the-Loop     & Feedback Loops, RLHF for Data \\
Domain-Specific Generators & Medical, Legal, Multilingual \\
Benchmarking          & Standard Datasets, Competitions, Reproducible Pipelines \\
\hline
\end{tabular}
\end{table*}

\section{Conclusion}

Synthetic data generation using large language models has emerged as a powerful technique to address data scarcity and enhance model training in both natural language and programming language domains. We have reviewed the state-of-the-art advances in this area, highlighting how LLMs can generate high-quality text and code data that, when used judiciously, lead to significant performance gains on downstream tasks. 

In the \textbf{text domain}, LLM-generated data has proven especially useful for low-resource scenarios, delivering substantial improvements in tasks like classification and QA when human-labeled data is limited \cite{li_data_2024}. Techniques such as prompt-based augmentation (zero-shot, few-shot, etc.), retrieval augmentation for grounding facts \cite{chai_text_2025}, and iterative refinement have pushed synthetic text data closer in effectiveness to real data. 

In the \textbf{code domain}, LLMs have unlocked new possibilities by generating code snippets, programming instructions, and even whole problem solutions, facilitating better training of code models. The ability to verify code correctness via execution \cite{liu_best_2024} provides a strong advantage, allowing the curation of large-scale, correct-by-construction synthetic code datasets (e.g., Code Alpaca, WizardCoder) that have propelled open-source code models to approach the competency of their proprietary counterparts.

Throughout this survey, we have also underscored the \textbf{practical considerations}: synthetic data must be used with care to ensure quality and avoid pitfalls like distribution shift, bias amplification, or model collapse from feedback loops. Encouraging findings, such as the avoidance of model collapse by combining synthetic with real data \cite{gerstgrasser_is_2024}, give confidence that these pitfalls can be managed with thoughtful strategies. The community has developed numerous methods (from filtering heuristics to advanced RL techniques) to maximize the signal and minimize the noise in synthetic datasets \cite{li_data_2024}.

Our discussion on challenges reflects that this field is still evolving. Issues of factual accuracy, diversity, and evaluation metrics remain open for continued research. However, the rapid progress in just the last two years — with surveys cataloging dozens of new works \cite{chen_mastering_2025} — indicates a vibrant research momentum. We are likely to see even more innovative solutions, like integrated human-in-the-loop generation and domain-specific synthetic data models, being realized soon.

In conclusion, LLM-based synthetic data generation stands as a \textbf{promising paradigm shift} in how we obtain training data for AI. By leveraging the generative power of advanced models, we reduce our reliance on large hand-curated datasets and open up opportunities to develop AI systems in domains and languages that suffer from data scarcity. This democratizes the development of AI, enabling customization and improvement even when real data is lacking. 

As we refine these techniques, backed by rigorous evaluations and ethical safeguards, synthetic data generation will become an increasingly standard part of the machine learning toolkit—complementing real data, accelerating development, and perhaps one day even supplanting the need for certain kinds of manual data collection altogether. 

The synergy between learning from data and generating data to learn from encapsulates a fascinating frontier for AI research, one that blurs the line between model and data and pushes us toward more autonomous, self-improving AI systems.

\bibliographystyle{plainnat}

\end{document}